\documentclass[review,3p,times,11pt,authoryear]{elsarticle}

\usepackage{enumitem}
\usepackage{lineno}
\usepackage[colorlinks,linkcolor=red,anchorcolor=blue,citecolor=green]{hyperref}
\usepackage{multirow}
\usepackage{siunitx}
\usepackage{rotating}
\usepackage{epstopdf}
\usepackage{graphicx}
\usepackage{subfigure}

\usepackage{epstopdf}
\usepackage{times}
\usepackage{longtable}

\usepackage{makeidx}
\usepackage{multirow}
\usepackage{multicol}
\usepackage[dvipsnames,svgnames,table]{xcolor}
\usepackage{graphicx}
\usepackage{epstopdf}
\usepackage{ulem}
\usepackage{url}

\usepackage{array}
\usepackage{amsmath,amssymb,amsfonts}
\usepackage{algorithmic}
\usepackage{algorithm}
\usepackage{graphicx}
\usepackage{textcomp}

\usepackage{epsfig}
\usepackage{makecell}
\usepackage{longtable,booktabs}
\usepackage{longtable}

\usepackage{lscape}

\usepackage[export]{adjustbox}

\usepackage{natbib}
\usepackage[skip=0.333\baselineskip]{caption}

\usepackage{amsmath}

\journal{Preprint}
\begin{document}

\begin{frontmatter}

\title{Knowledge-Data Dually Driven Paradigm for Accurate Landslide Susceptibility Prediction under Data-Scarce Conditions Using Geomorphic Priors and Tabular Foundation Model}

\author[label1]{Yuting Yang}
\author[label1]{Gang Mei\corref{cor1}}
\ead{gang.mei@cugb.edu.cn}
\cortext[cor1]{Corresponding author}
\author[label1]{Feng Chen}
\author[label1]{Yongshuang Zhang}
\author[label1,label2]{Jianbing Peng}

\address[label1]{School of Engineering and Technology, China University of Geosciences (Beijing), 100083, Beijing, China}
\address[label2]{School of Geological Engineering and Geomatics, Chang'an University, Xi'an, 710064, China}


\begin{abstract}
Landslide susceptibility prediction is critical for geohazard risk assessment and mitigation. Conventional data-driven paradigm achieves high predictive accuracy but require sufficient conditioning factors and large-scale landslide inventories. However, in practical engineering applications across mountainous and plateau regions, data-scarce conditions are commonly observed, where such data requirements are rarely satisfied, rendering conventional data-driven paradigm inapplicable. To address this issue, we propose a knowledge-data dually driven paradigm for accurate landslide susceptibility prediction under data-scarce conditions. The essential idea behind the proposed novel paradigm is the integration of the geomorphic prior knowledge with scarce landslide data. To validate the proposed paradigm, we first applied it to a data-rich region in central Italy, where a conventional data-driven paradigm trained on the full dataset served as the baseline. By utilizing only 30\% of the available landslide data, the proposed paradigm achieved comparable predictive accuracy to the baseline, demonstrating its effectiveness under data-scarce conditions. The paradigm was further evaluated in a genuinely data-scarce environment for application, the Qilian Permafrost Region of the Tibetan Plateau, where it also yielded reliable susceptibility predictions, confirming its applicability under data-scarce conditions.
\end{abstract}

\end{frontmatter}

\section{Introduction}

Landslides rank among the most frequent and destructive geological hazards worldwide \citep{RN1}, driven by the complex interplay of topographic, geological, hydroclimatic, and anthropogenic factors \citep{RN2,RN3}. As extreme climatic events intensify and human encroachment into vulnerable montane environments accelerates, both the frequency and severity of landslide occurrence are increasing globally \citep{RN4,RN5}. Between 2004 and 2016 alone, landslides claimed more than 55,000 lives and inflicted annual economic losses exceeding USD 20 billion, with no indication of abatement \citep{RN6,RN7}. Against this backdrop, Landslide Susceptibility Mapping (LSM) has emerged as an indispensable component of geohazard risk assessment, providing spatially explicit identification of landslide-prone terrain to inform land-use planning, disaster mitigation, and emergency management \citep{RN8}. Developing accurate, computationally efficient, and broadly transferable susceptibility prediction paradigms therefore constitutes a central scientific challenge in contemporary geohazard research.

The rapid advancement of geospatial technologies and artificial intelligence has established conventional data-driven paradigm as the dominant paradigm for landslide susceptibility prediction \citep{RN9,RN10}. Extensive evidence indicates that within this data-driven paradigm, algorithms ranging from conventional statistical models to conventional machine learning models can effectively capture the complex nonlinear relationships between landslide occurrence and multi-source environmental predictors, providing accurate landslide susceptibility predictions \citep{RN11,RN12}. However, this purely data-driven paradigm, is critically contingent on two prerequisites: comprehensive, high-quality landslide inventory data and a rich set of multi-dimensional conditioning factors (e.g., topography, geology, land use, and climate). Consequently, the predictive performance of this paradigm is highly sensitive to both the completeness and the accuracy of the available data \citep{RN13}.

In practice, these prerequisites are rarely satisfied simultaneously. Historical landslide inventories are difficult to compile, but the demand for reliable susceptibility maps is greatest precisely in regions where such inventories are most lacking \citep{RN14}. Scarce landslide inventories severely degrade overall predictive capacity, posing a challenge that existing conventional data-driven paradigm cannot easily overcome \citep{RN15,RN16}. The conventional data-driven paradigm, when trained on small datasets, is prone to significant declines in predictive accuracy, with performance further deteriorating under the complex geological conditions characteristic of data-scarce regions \citep{RN17}. These challenges are particularly acute in geologically and ecologically sensitive environments such as the Tibetan Plateau, where permafrost degradation and freeze--thaw cycling intensify slope instability, while the acquisition of geological, lithological, and land-use data faces severe constraints \citep{RN18}. The scarce data severely constrains the applicability and generalizability of this paradigm, and may even lead to pronounced bias in susceptibility estimates \citep{RN19}. How to achieve reliable landslide susceptibility prediction under data-scarce conditions has therefore emerged as a critical problem.

To address this problem, we propose a novel knowledge-data dually driven paradigm for landslide susceptibility prediction under data-scarce conditions. The essential idea behind the proposed novel paradigm is the integration of the geomorphic prior knowledge with scarce landslide data. Specifically, our paradigm first applies the morphometric model to estimate the geomorphic prior knowledge. Subsequently, these geomorphic priors are integrated with conventional predictive factors and landslide inventory to construct an enhanced, prior-constrained dataset. Finally, a foundation model designed for small data prediction (e.g., TabPFN) is applied for predicting, thereby achieving high-precision predictions under data-scarce conditions. The proposed paradigm maintains the geomorphic prior knowledge while leveraging the advantages of the foundation model in handling scarce data. This improves the generalization and stability of the paradigm, providing a highly practical solution for landslide susceptibility prediction in data-scarce and complex geological environments, such as the Qinghai-Tibet Plateau.

\section{Methods: Proposed paradigm for landslide susceptibility prediction in data-scarce regions}

\subsection{Overview}

This study proposes a novel knowledge-data dually driven paradigm for landslide susceptibility prediction under data-scarce conditions. The essential idea behind the proposed novel paradigm is the integration of the geomorphic prior knowledge with scarce landslide data. Under conditions of data-scarce, the conventional data-driven paradigm is highly susceptible to overfitting, as they tend to capture spurious correlations inherent in small sample sets. To address this problem, the proposed paradigm incorporates geomorphic prior knowledge as a physical constraint, ensuring that the susceptibility assessments consistently adhere to basic topographic and mechanical principles. Concurrently, recognizing that sparse landslide datasets harbor intrinsic geomorphic regularities despite their limited volume, a specialized foundation model is employed as a targeted analytical tool. Its architecture facilitates the distillation of these latent patterns, thereby ensuring robust inference. Through this synergistic mechanism of ``physical prior knowledge and data-driven inference,'' the proposed paradigm effectively mitigates the dual risks of statistical bias and geologically unrealistic predictions arising from data scarcity.

\begin{figure}[!ht]
\centering
\includegraphics[width=0.95\textwidth]{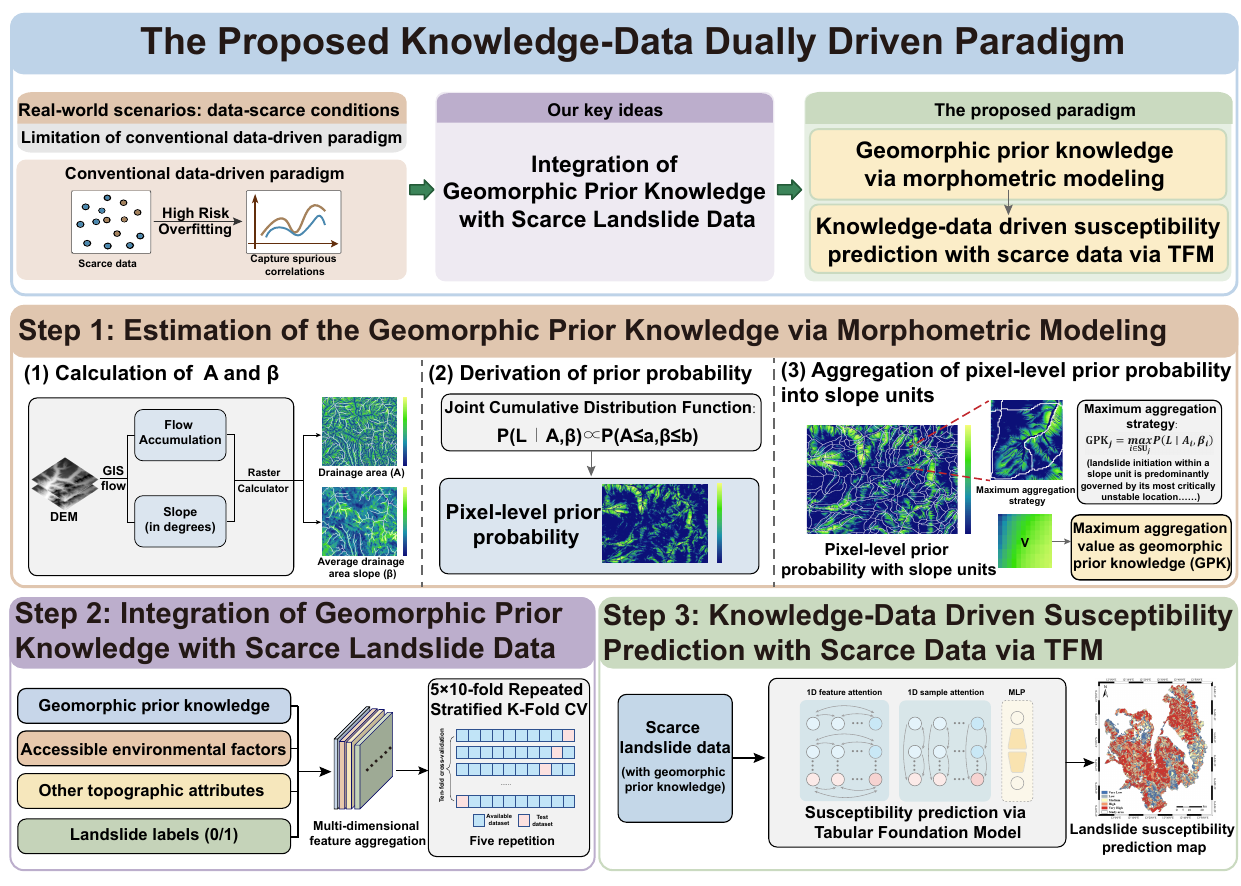}
\caption{The workflow of the proposed paradigm.}
\label{Figure1}
\end{figure}

For the implementation, the paradigm comprises three steps (Fig.\ref{Figure1}). Step 1: Estimation of the geomorphic prior knowledge is achieved via morphometric modeling. Specifically, our paradigm first applies the morphometric model to estimate the morphological stability of landslides in the study area using only DEM data. Instead of serving as the final susceptibility prediction, these calculated values act as geomorphic prior knowledge. Step 2: The geomorphic prior knowledge is integrated with the scarce landslide data. The geomorphic priors are integrated with conventional predictive factors and a scarce landslide inventory to construct an enhanced, prior-constrained dataset. Step 3: Knowledge-data dually driven landslide susceptibility prediction is executed under data-scarce conditions via a foundation model, thereby achieving high-precision predictions. The complete workflow is illustrated in Figure X. The detailed technical implementation of each step is described in Sections \ref{2.2} to \ref{2.4}.

\subsection{Step 1: Estimation of the geomorphic prior knowledge via morphometric modeling}
\label{2.2}

\subsubsection{Calculation of drainage area and average drainage area slope}

The morphometric model requires two primary topographic inputs: upstream drainage area (A) and average drainage area slope ($\beta$). Both factors were derived within ArcGIS using the D8 flow-direction algorithm. Raw DEM data were first preprocessed to generate a hydrologically conditioned DEM (HydroDEM). Slope (S, in degrees) was computed from the original DEM using the ArcGIS Slope tool. Flow Direction (D8 algorithm) and Flow Accumulation operations were then applied sequentially to the HydroDEM, yielding a per-pixel flow accumulation raster ($F_{a c c}$). The drainage area was expressed in natural logarithmic form as given in Equation \ref{eq1}:

\begin{equation}
    \label{eq1}
	A=\ln \left[\left(F_{a c c}+1\right) \times r^2\right]
\end{equation}
where r² denotes the area of a single grid cell (m²).

The average drainage area slope, $\beta$, represents the area-weighted mean slope of the entire upstream contributing area for a given pixel, computed as shown in Equation \ref{eq2}:

\begin{equation}
	\label{eq2}
	\beta=\frac{F_{a c c}^S+S}{F_{a c c}+1}
\end{equation}
where ($F_{a c c}$ is the weighted flow accumulation value computed with S as the weight raster, obtained by setting the Weight Raster parameter to S in the ArcGIS Flow Accumulation function.

Subsequently, stream network rasters were generated for each study area. Buffer zones were established on both sides of the delineated channels, and the Con function was applied to assign NoData values to A and $\beta$ within these buffers, thereby eliminating the influence of channelized flow on model performance.

\subsubsection{Derivation of pixel-level prior probability via joint cumulative distribution function}
\label{2.2.2}

The joint cumulative distribution function (JCDF) of A and $\beta$ was computed across the study area as a per-pixel estimate of shallow landslide prior susceptibility probability, following the core principles of the morphometric model proposed by \cite{RN14}, as expressed in Equation \ref{eq3}:

\begin{equation}
	\label{eq3}
P(L\mid A,\beta)\propto P(A\leq a,\beta\leq b)
\end{equation}
where $\beta$ is the average drainage area slope ($^\circ$) and A is the drainage area (m²). The left-hand side represents the conditional probability of landslide occurrence given A and $\beta$; the right-hand side is the corresponding JCDF, denoting the probability that both A and $\beta$ within the study domain simultaneously fall below specified thresholds a and b. The physical interpretation is straightforward: pixels whose slope and contributing area rank closer to the upper tail of the study-domain distribution are assigned higher prior probabilities of shallow landslide occurrence.

It should be noted that the original formulation employs a multi-scale hierarchical modeling strategy, computing the JCDF separately within nested watershed units before integrating results through area-weighted aggregation. In this study, the procedure is simplified by computing a single JCDF over the entire study area, thereby reducing implementation complexity while ensuring methodological consistency across both study areas.

Prior probability estimation was performed only for pixels exceeding a specified slope filtering threshold; pixels below this threshold were assigned a prior susceptibility of zero. Similarly, pixels masked within channel buffer zones, along with other missing data (NoData) regions, were uniformly assigned a value of zero to ensure spatially continuous coverage of the prior susceptibility field. The specific values for these two key morphometric parameters---the slope threshold and the channel buffer width---were determined via a domain-wide sensitivity analysis. By simultaneously maximizing the Area Under the Curve (AUC) and minimizing the Brier Score of the morphometric outputs against ground-truth landslide inventories, the optimal configurations were identified: a 10$^\circ$ slope threshold and 40 m buffer for the central Italy study area, and a 20$^\circ$ slope threshold and 60 m buffer for the Qilian Permafrost Region. Additionally, morphological processing was applied to enforce topological closure along study area boundaries, enabling precise differentiation between internal no-data regions and external background areas, and thereby preventing boundary-induced mapping errors.

\subsubsection{Aggregation of pixel-level prior probability into slope units}

As slope units (SUs) serve as the fundamental mapping units in this study, the pixel-scale outputs of the morphometric model must be aggregated to the SU scale. Following the standard processing protocol of \cite{RN14}, a maximum aggregation strategy is adopted, whereby the geomorphic prior knowledge (GPK) for each SU is defined as the maximum prior susceptibility value among all valid pixels within that unit:

\begin{equation}
	\label{eq4}
	\mathrm{GPK}_j=\max _{i \in \mathrm{SU}_j} P\left(L \mid A_i, \beta_i\right)
\end{equation}
where $j$ is the SU index; $SU_{j}$ denotes the j-th slope unit; i indexes valid pixels within $SU_{j}$; and $P\left(L\mid A_{i},\beta_{i}\right)$ is the pixel-level landslide prior probability estimated by the morphometric model, with $A_{i}$ and $\beta_{i}$ representing the contributing area and slope gradient of pixel $i$, respectively.

The rationale for selecting the maximum rather than the mean is as follows: landslide initiation within a slope unit is predominantly governed by its most critically unstable location, and mean aggregation would underestimate true unit-level susceptibility. Within the proposed paradigm, geomorphic prior knowledge is not used directly to generate the final susceptibility map; rather, it serves as a physically informed prior feature that encodes terrain-mechanical information and is incorporated into the input feature space of the subsequent tabular foundation model.

\subsection{Step 2: Integration of geomorphic prior knowledge with scarce landslide data}

Based on the scarce landslide inventory, geomorphic prior knowledge is combined with additional topographic attributes (slope gradient, aspect, etc.), and accessible environmental factors to form the input feature. The incorporation of geomorphic prior knowledge transforms the slope stability mechanisms inherent in terrain morphology into quantifiable conditioning factors. This provides a spatially continuous geomorphic context for susceptibility mapping in regions where landslide inventories are scarce, effectively compensating for the spatial sparsity and inherent sampling biases typical of existing landslide records.

\subsection{Step 3: Knowledge-data dually driven landslide susceptibility prediction under data-scarce conditions via foundation model}
\label{2.4}

In this study, tabular foundation model \citep{RN20} is adopted as the specific implementation of the foundation model. Within our proposed paradigm, the tabular foundation model receives the multivariable conditioning factors that are crucially anchored by the extracted geomorphic prior knowledge. By utilizing a context-driven inference mechanism instead of exhaustive local parameter training, the tabular foundation model identifies the underlying geomechanically mechanisms governing slope instability and distills the latent regularities inherent within the scarce data itself, effectively preventing overfitting.

Within the proposed paradigm, the tabular foundation model receives conditioning factors incorporating geomorphic prior knowledge, is fitted on labeled SU samples (landslide/non-landslide), and outputs landslide occurrence probabilities $\hat{p}_{j}$$\in$[0, 1] for all slope units:

\begin{equation}
\hat{p}_{j}=f_{TFM}\left(x_{j}^{(SU)}\mid D_{train }^{(SU)}\right)
\end{equation}
where $x_{j}^{(SU)}$ represents the comprehensive conditioning factors of the j-th slope unit (integrating the physics-based geomorphic prior knowledge alongside supplementary topographic and environmental factors), and $D_{train }^{(SU)}$ denotes the available scarce training dataset.

Through this profound integration of geomorphic prior knowledge and foundation model inference, the proposed paradigm ensures highly reliable and geologically sound predictions even under data-scarce conditions.

\section{Verification}

\subsection{Data collection and preprocessing}

\begin{figure}[!ht]
\centering
\includegraphics[width=0.95\textwidth]{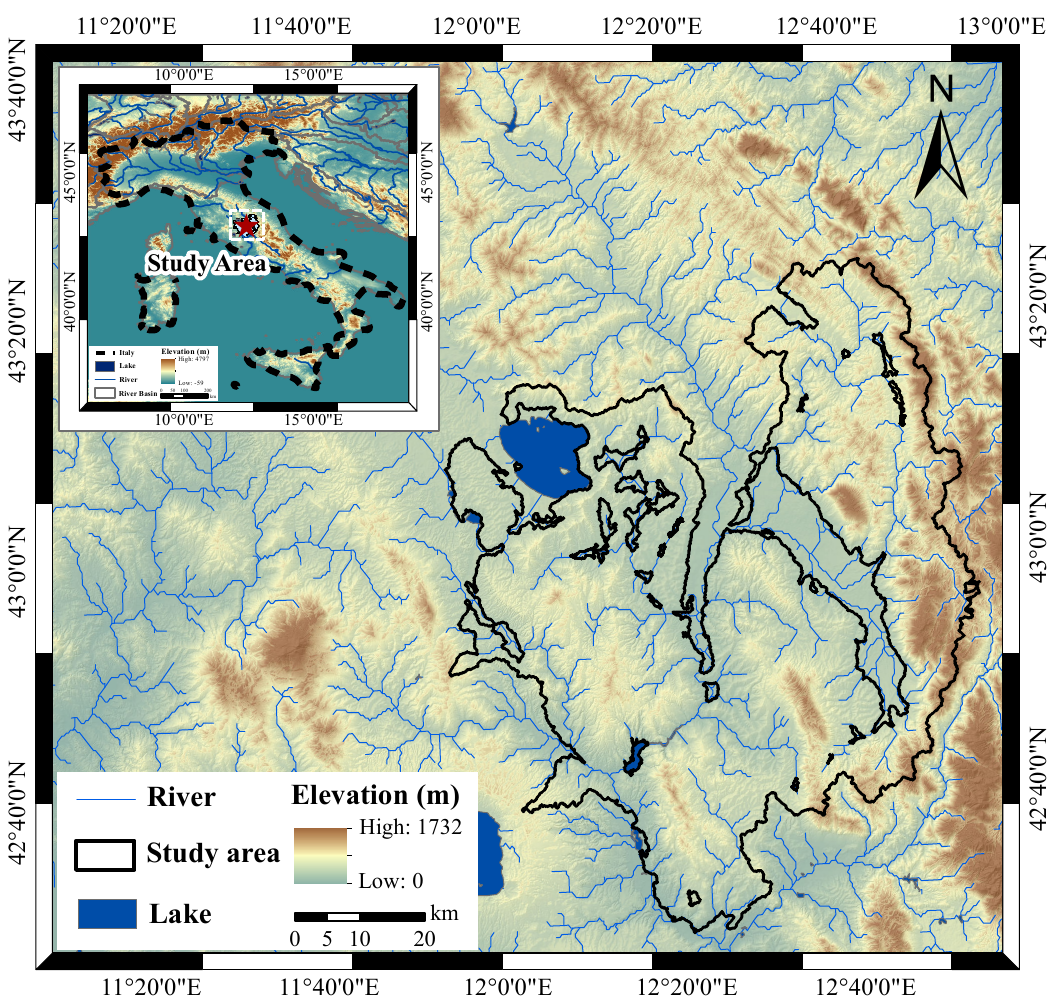}
\caption{Regional location of the Central Italy study area.}
\label{Figure2}
\end{figure}

We utilize the landslide susceptibility benchmark dataset developed by \cite{RN21}. It is located in Central Italy, encompassing an area of approximately 4,100 $km^2$ \citep{RN21}. This region serves as a standardized benchmark environment characterized by well-documented translational landslides. The robust geological and environmental records available for this region provide a baseline to verify the paradigm's performance. The location of the study area is illustrated in Fig.\ref{Figure2}.

\begin{figure}[!ht]
\centering
\includegraphics[width=0.95\textwidth]{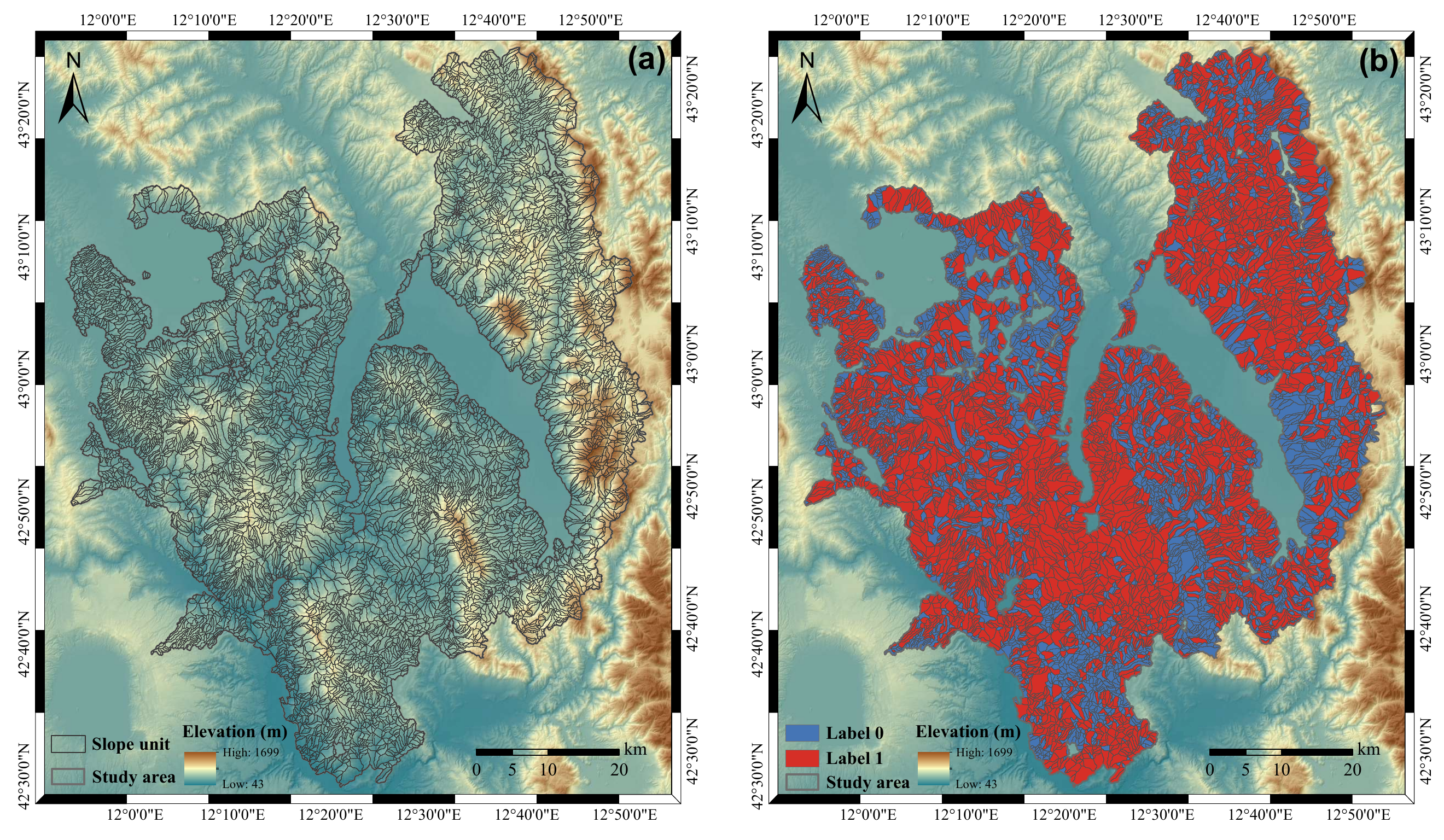}
\caption{delineation of Slope unit and distribution of labels in the Central Italy study area.}
\label{Figure3}
\end{figure}

This benchmark provides SUs pre-delineated using high-resolution topographic data, enabling direct application without secondary spatial partitioning. To construct the target labels, we extracted the attribute table from the official vector layer, which records two distinct landslide occurrence scenarios (p1 and p2). Scenario p1 was selected as the supervision label for the binary classification model. Following rigorous screening and balancing, the dataset comprises 7,360 SUs with a nearly 1:1 ratio of landslide (Label = 1; 3,594 SUs) to non-landslide (Label = 0; 3,766 SUs) instances. The distribution of slope units and label distribution in the study area are illustrated in Fig.\ref{Figure3}.

Following the rigorous feature selection and principal component optimization established in the benchmark study, 19 predictive factors were retained to model susceptibility. These encompass: Morphometric variables: mean slope steepness, mean profile curvature, mean and standard deviation of northerness and easterness, and the maximum distance divided by the square root of the SU area. Hydrological indices: standard deviation of the topographic wetness index. Soil properties: mean and standard deviation of bulk density, mean weight percentage of clay, sand, and silt particles, and the standard deviation of silt weight percentage. Lithological variables: areal coverage percentages of alluvial deposits, unconsolidated sedimentary rocks, marlstone, schistose metamorphic rocks, and carbonate rocks.

\subsection{Experimental design for verification}

To investigate the learning curve and generalization capacity of the proposed paradigm across these varying conditions---ranging from extreme data scarcity to relative data sufficiency---a multi-gradient sample size scaling experiment was designed. Within the established training dataset (available data), a series of incrementally increasing sample sizes was constructed for the study area. Stratified sampling was strictly enforced at every sample size level to preserve the original class distribution. By tracking the evolution of model performance metrics (Area Under the Receiver Operating Characteristic Curve [AUC-ROC] and Brier Score [BS]) across increasing data volumes, the incremental predictive benefit of the proposed paradigm for landslide susceptibility mapping in data-scarce regions is quantitatively characterized.

To ensure the statistical robustness of performance estimates, model evaluation was conducted using a 5×10-fold Repeated Stratified K-Fold Cross-Validation scheme, yielding 50 independent training and evaluation cycles in total. In each cycle, all slope unit (SU) samples were randomly partitioned into a training set (available data, 90\%) and a held-out testing set (unavailable data, 10\%). Stratified sampling was strictly enforced to maintain the positive-to-negative class ratio in both subsets consistent with that of the full dataset. Furthermore, to provide an objective assessment of the relative gains offered by the proposed paradigm, conventional data-driven baselines (Random Forest \citep{RN22} and XGBoost \citep{RN23}) were trained under identical data partitioning conditions. These algorithms were selected as they represent the most extensively validated conventional paradigms in current landslide susceptibility literature \citep{RN24,RN25}. Detailed configurations for these baselines, alongside the rigorous mathematical formulations and justification for the complementary AUC-ROC \citep{RN26} and Brier Score \citep{RN27} evaluation metrics, are provided in the Supporting Information A. The five repetitions effectively suppress variance attributable to any single random partition, rendering the reported performance metrics (AUC and Brier Score) highly objective and reproducible, and ultimately translating into susceptibility maps with smoother, more reliable spatial transitions.

\subsection{Results and analysis of verification}

\subsubsection{Comparative results of landslides susceptibility prediction}

To visually assess the spatial generalization capability and predictive plausibility of different paradigms at a regional scale, we generated comprehensive landslide susceptibility maps. We employed an ensemble averaging strategy for the final global mapping to maximize the utility of the geo-environmental information embedded within the available samples. Specifically, the 50 sub-models derived from the cross-validation iterations were applied to all slope units across the study area. Their predicted probabilities were then averaged to yield the final Landslide Susceptibility Index (LSI) for each unit. By aggregating the outputs of 50 models trained on slightly perturbed datasets, this approach effectively mitigates the spatial prediction variance inherent to single-training runs. Consequently, the resulting susceptibility maps exhibit smoother, more reliable spatial transitions that better align with actual geomorphic patterns.

Furthermore, for the multi-paradigm spatial comparison, applying independent classification thresholds would distort the visual baseline due to the stark disparities in probability calibration characteristics among different paradigms. To resolve this, we implemented a unified ``Baseline Projection'' classification strategy. Specifically, a standardized hazard baseline was established using the Jenks Natural Breaks method based on our paradigm's optimal full-sample predictions, yielding thresholds of 0.22, 0.41, 0.59, and 0.77 for the five susceptibility levels (from very low to very high). This baseline classification was uniformly applied to the prediction outputs of the conventional data-driven paradigm (XGBoost and Random Forest), ensuring an unbiased spatial cross-comparison.

\begin{figure}[!ht]
\centering
\includegraphics[width=0.95\textwidth]{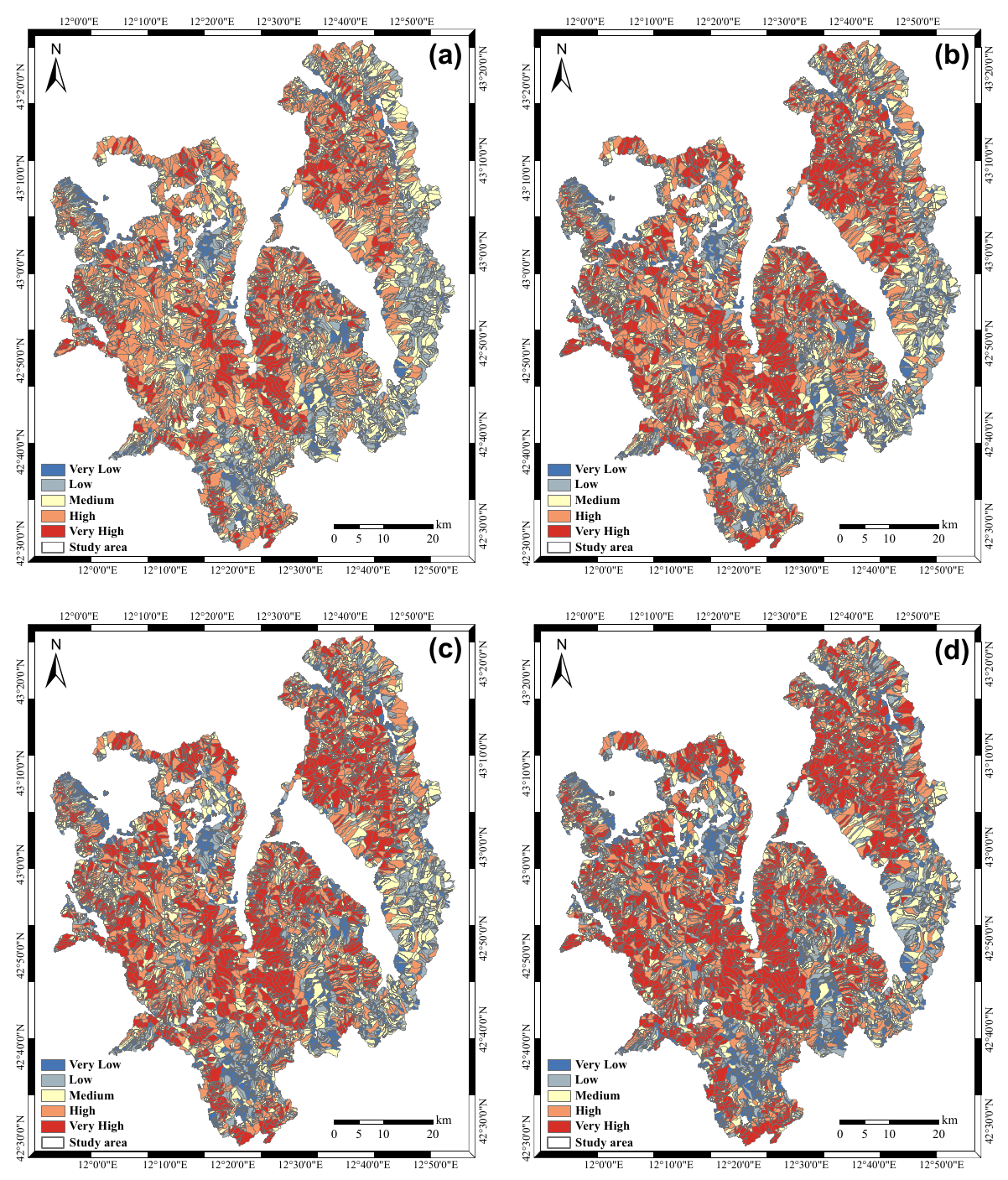}
\caption{Landslide susceptibility maps of the central Italy study area via the proposed paradigm: (a) using 10\% available landslide data; (b) using 30\% available landslide data; (c) using 50\% available landslide data; (d) using 100\% available landslide data.}
\label{Figure4}
\end{figure}

\begin{figure}[!ht]
\centering
\includegraphics[width=0.95\textwidth]{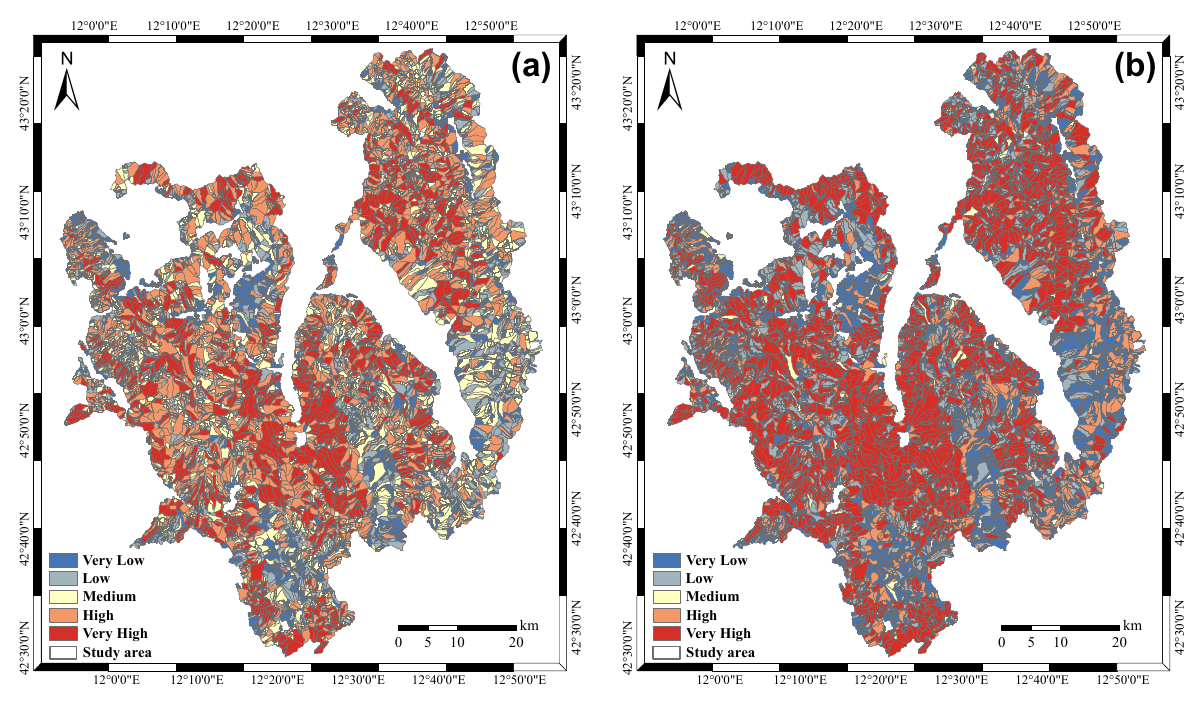}
\caption{Landslide susceptibility maps of the central Italy study area via conventional data-driven paradigm: (a) conventional data-driven paradigm (XGBoost using 100\% available landslide data); (b) conventional data-driven paradigm (RF using 100\% available landslide data).}
\label{Figure5}
\end{figure}

Fig.\ref{Figure4} presents the landslide susceptibility maps for the central Italy study area generated by the proposed paradigm across varying levels of available landslide data (10\%, 30\%, 50\%, and 100\%). Fig.\ref{Figure5} illustrates the corresponding maps generated by the conventional data-driven paradigm (XGBoost and RF) utilizing 100\% of the available data.

A primary advantage demonstrated by the proposed paradigm is its exceptional robustness in data-scarce scenarios. As observed in Fig.\ref{Figure4}a, even when constrained to a mere 10\% of the available landslide inventory, the model effectively delineates the fundamental geomorphological susceptibility trends. The spatial distribution patterns remain highly consistent and geomorphologically coherent as the training data volume progressively increases from 10\% to 100\% (Fig.\ref{Figure4}a--d). This remarkable stability indicates that the model is not merely memorizing sample locations. By integrating physical geomorphic priors with a Tabular Foundation Model (TFM)---whose architecture inherently facilitates the distillation of underlying spatial patterns rather than being explicitly designed to do so---the proposed paradigm successfully mitigates the overfitting typically associated with small-sample datasets.

Regarding spatial distribution characteristics at the full data threshold (Fig.\ref{Figure4}d), our paradigm demonstrates excellent spatial continuity and geomorphic coherence. The identified very high (red) and high (orange) susceptibility zones display a significant contiguous, belt-like distribution (e.g., the coherent high-risk bands in the eastern and central parts of the study area). This smooth probabilistic transition effectively prevents abrupt spatial discontinuities, proving that under the constraint of geomorphic priors, the model captures the slope instability patterns governed by macroscopic geological structures and main drainage systems.

In contrast, while both purely data-driven models of the conventional paradigm (Fig.\ref{Figure5}) captured certain baseline trends, they exposed distinct limitations in spatial generalization within complex geological settings, leaving distinctly different algorithmic signatures in their mapped outputs.

Among them, Random Forest (RF) (Fig.\ref{Figure5}b) suffered from the most severe spatial overfitting. Evaluated against the unified benchmark thresholds, the RF predictions displayed an extreme "salt-and-pepper effect." Its map is dominated by isolated, fragmented red patches, with frequent and abrupt adjacencies between very high (dark red) and very low (dark blue) susceptibility zones, largely failing to produce reasonable moderate-risk (yellow/orange) transition areas. This indicates that RF over-relied on local training sample features, resulting in highly fragmented spatial predictions that fundamentally contradict the natural, gradual transitions typical of geological and environmental factors.

Conversely, benefiting from its internal gradient boosting mechanism and regularization penalties, XGBoost (Fig.\ref{Figure5}a) effectively mitigated the extreme polarization observed in RF. As the spatial distribution map illustrates, XGBoost reasonably captured macroscopic susceptibility trends; its identified "high susceptibility zones" (orange) share a broad morphological resemblance with those of our paradigm. However, when delineating the most critical "very high susceptibility zones" (red), the purely data-driven nature of XGBoost still revealed the drawbacks of lacking geomorphic constraints. Its highest risk zones exhibit significant fragmentation, scattered irregularly across the orange background. In contrast, our paradigm precisely aggregates these very high-risk zones into coherent belts along key geomorphic structural controls, such as main river valleys and tectonic fault zones.

\subsubsection{Comparative analysis of landslides susceptibility prediction}

\begin{figure}[!ht]
\centering
\includegraphics[width=0.95\textwidth]{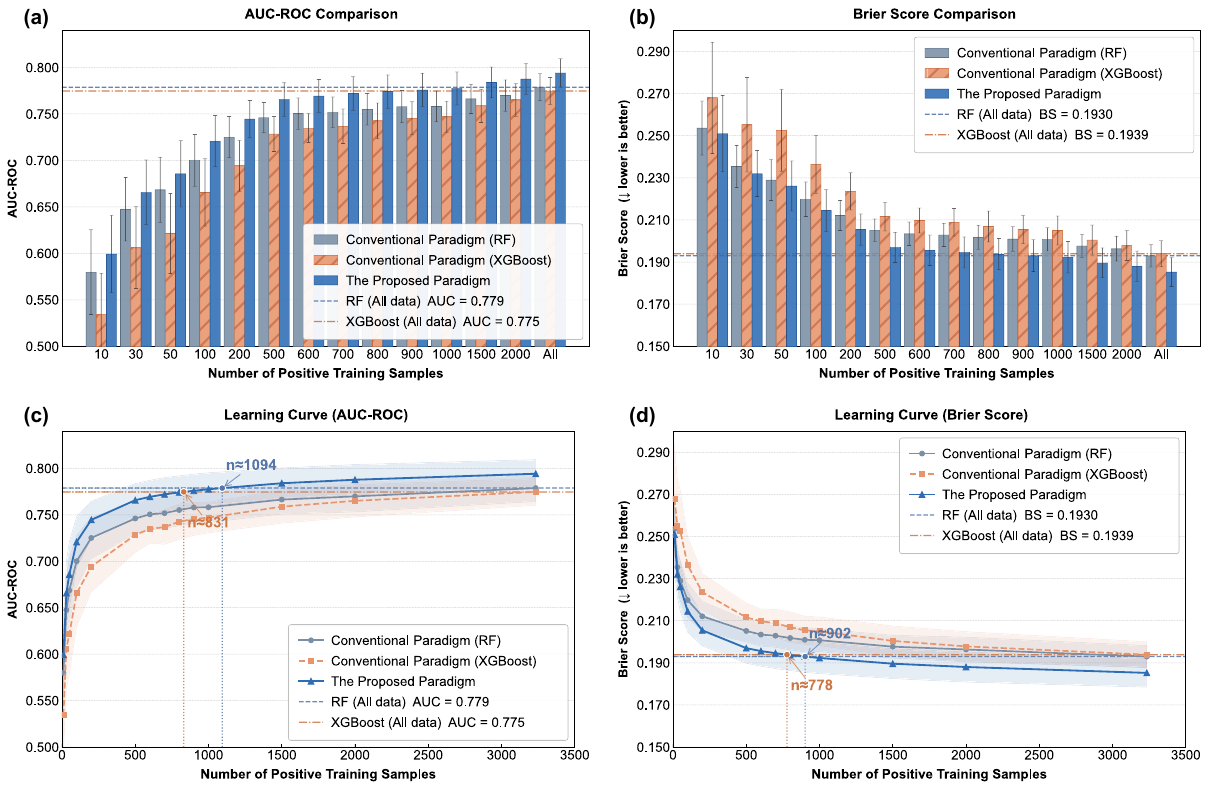}
\caption{Comparison of predictive performance between the proposed paradigm and conventional data-driven paradigm under varying levels of available samples: (a) AUC comparison; (b) BS comparison; (c) AUC learning curves; (d) BS learning curves.}
\label{Figure6}
\end{figure}

Under conditions of a 1:1 positive-to-negative sample ratio and the full set of environmental and topographic covariates, the predictive performance of the proposed paradigm was compared against that of the conventional data-driven paradigm (RF and XGBoost) across a range of training sample sizes. Fig.\ref{Figure6} presents the AUC and Brier Score learning curves for all paradigms as a function of sample size.

Overall, the proposed paradigm consistently outperforms the conventional data-driven baseline across all sample sizes examined. This advantage is most pronounced under data-scarce conditions (number of positive samples $\leq$ 100, corresponding to a total sample count of $\leq$ 204), where the proposed paradigm achieves substantially higher AUC and lower Brier Score, demonstrating markedly superior discriminative capacity and probabilistic calibration under data-limited conditions.

As training sample size increases, performance improves across all paradigms. However, the conventional data-driven paradigm exhibits a steeper improvement trajectory, progressively narrowing the performance gap with the proposed paradigm. This pattern indicates that the conventional data-driven paradigm is more heavily dependent on data volume, whereas the proposed paradigm---by virtue of its incorporated geomorphic prior knowledge---is capable of extracting meaningful predictive information even under conditions of severe data scarcity.

Notably, when the available sample size reaches approximately 30\% of the total available samples, the proposed paradigm already achieves or surpasses the predictive performance of the conventional data-driven baselines trained on the full available samples. This result demonstrates a substantial advantage in data utilization efficiency of the proposed paradigm.

\section{Application}

\subsection{Study area}

\begin{figure}[!ht]
\centering
\includegraphics[width=0.95\textwidth]{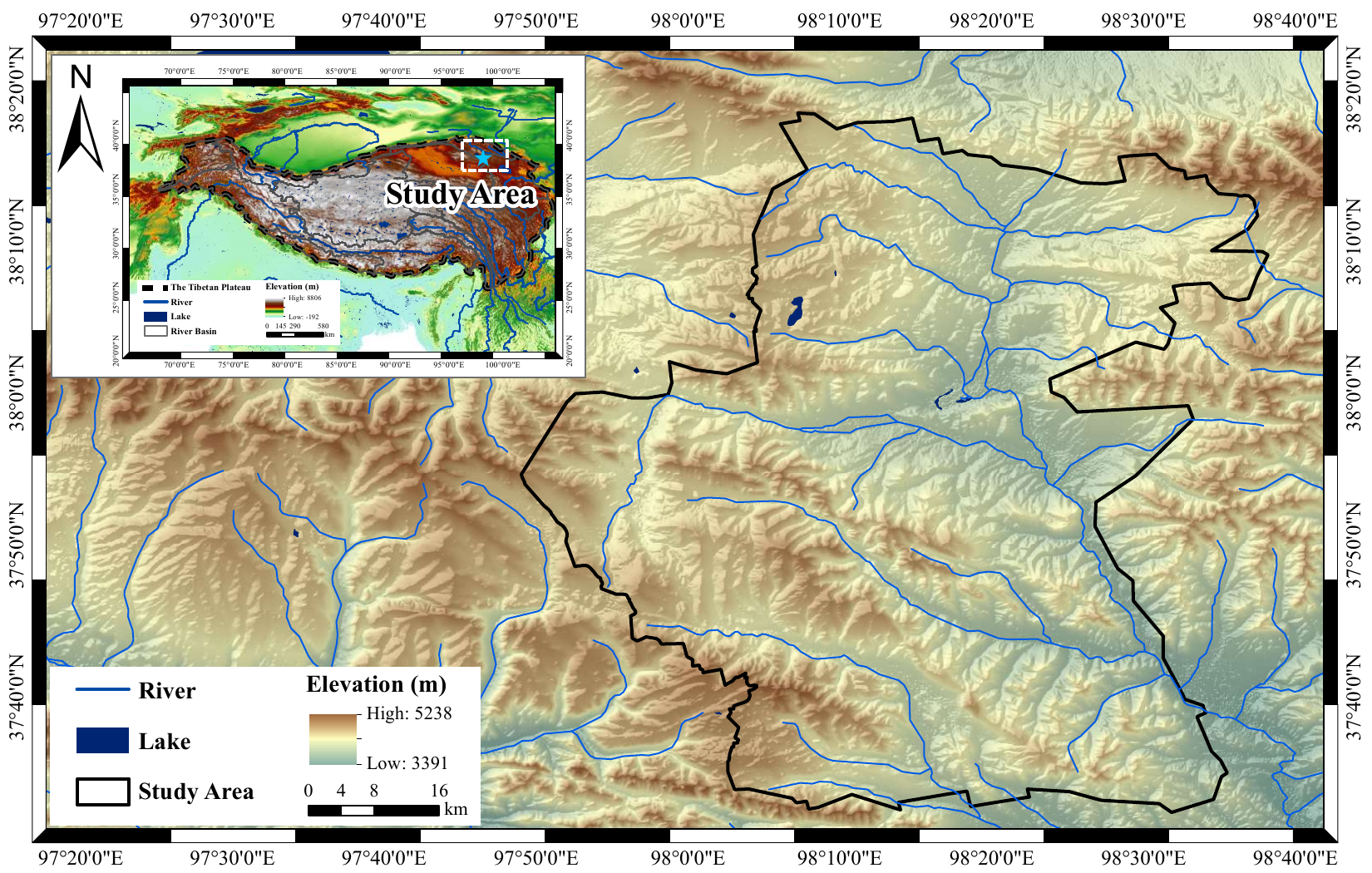}
\caption{Regional location of the Qilian Permafrost Region, Tibetan Plateau study area.}
\label{Figure7}
\end{figure}

The study area for application is located in the transition zone between the southern foothills of the Qilian Mountains and the northeastern margin of the Qaidam Basin, covering roughly 3,738 $km^2$ with elevations ranging from 3,657 to 4,952 m \citep{RN28}. It features a complex alpine permafrost environment where slope stability is profoundly governed by high-altitude freeze-thaw cycles, extreme topographic variations, and active neotectonics. Unlike the central Italy study area, this area suffers from scarce landslide inventory data, representing a typical "data-scarce" environment.The location of the study area is illustrated in Fig.\ref{Figure7}.

\subsection{Data collection and preprocessing}

The study area is situated in the Qilian Mountains permafrost region on the northeastern margin of the Tibetan Plateau \citep{RN28}. Landslide activity in this geologically complex terrain is governed by the coupled effects of alpine freeze-thaw cycles and extreme topography. Compared to the Central Italy dataset, landslide inventories here are notably sparser, typifying a ``data-scarce'' environment.

Because the original spatial data were not delineated by slope units, we applied the same morphological segmentation algorithm used in the central Italy study area benchmark dataset to ensure methodological consistency. SUs for the study area were delineated using the r.slopeunits module within GRASS GIS, driven by the regional digital elevation model (DEM).

\begin{figure}[!ht]
\centering
\includegraphics[width=0.95\textwidth]{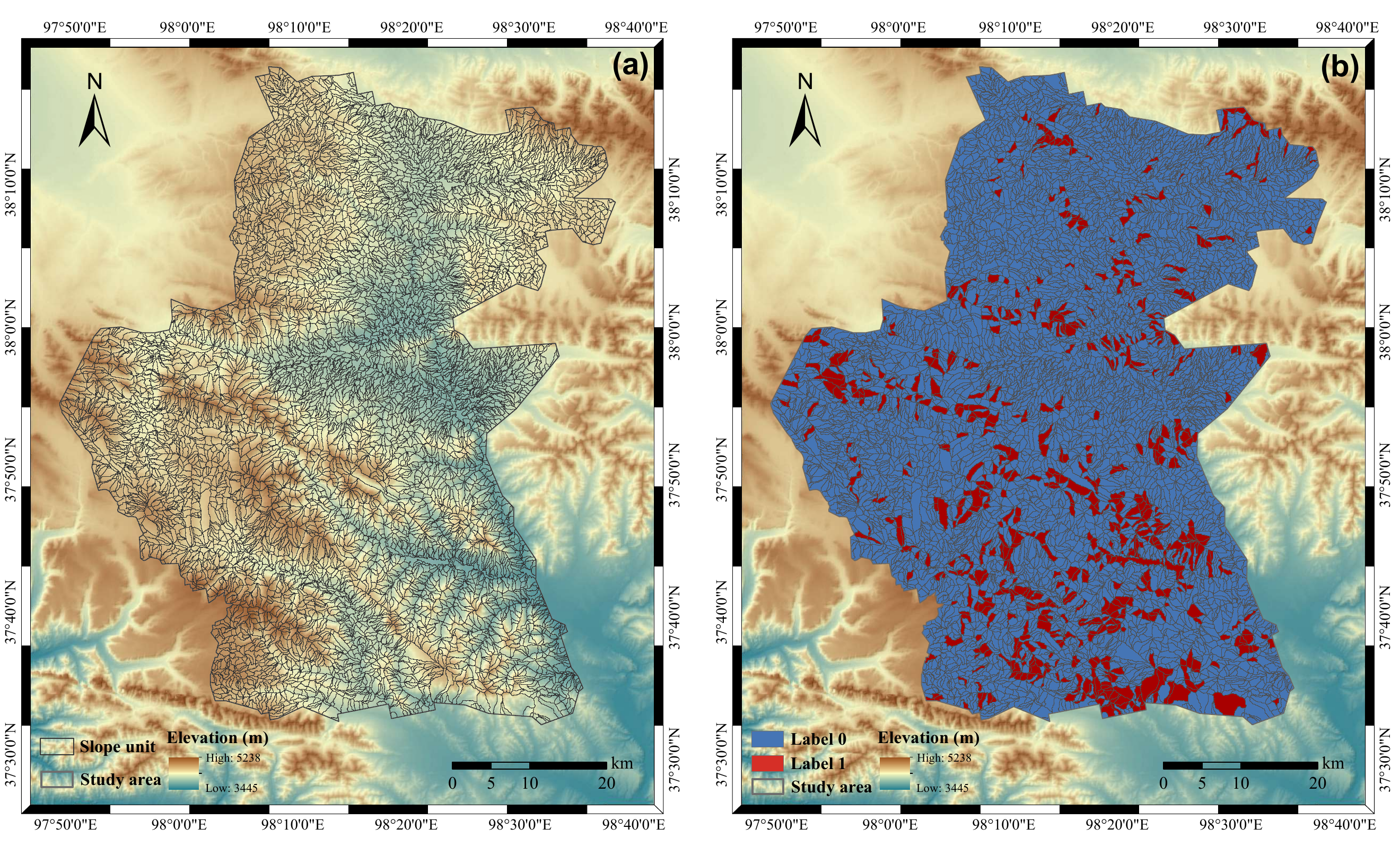}
\caption{delineation of Slope unit and distribution of labels in the Qilian Permafrost Region, Tibetan Plateau study area.}
\label{Figure8}
\end{figure}

Following this preprocessing workflow, the study area was partitioned into 6,145 valid SUs. A spatial join of the existing landslide inventory points with the SU layer yielded 654 landslide-bearing units (Label = 1) and 5,491 stable units (Label = 0). This ratio of approximately 1:8 represents a severe class imbalance. This highly challenging dataset is employed to rigorously evaluate the robustness and efficacy of our proposed paradigm under extreme conditions of data scarcity. The distribution of slope units and label distribution in the study area are illustrated in Fig.\ref{Figure8}.

Predictive factors retained for this region encompass: Topographic factors: mean elevation, mean slope gradient, circular mean aspect, and the sine (easterness) and cosine (northerness) components of aspect. Climatic factors: the annual average number of freeze-thaw cycles.

\subsection{Results and analysis of application}

\begin{figure}[!ht]
\centering
\includegraphics[width=0.95\textwidth]{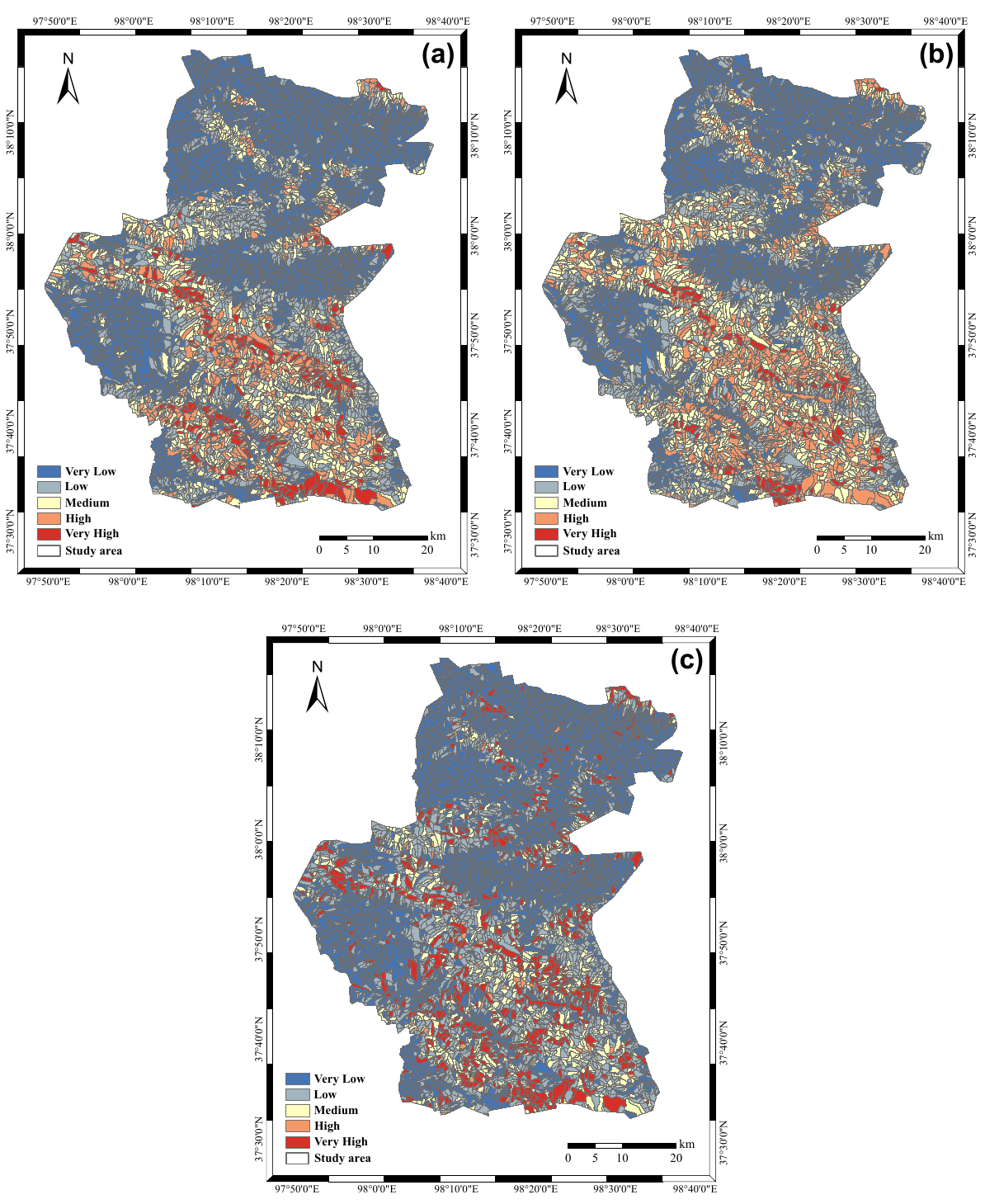}
\caption{Landslide susceptibility maps of the Qilian Permafrost Region, Tibetan Plateau study area under different paradigms: (a) the proposed paradigm; (b) conventional data-driven paradigm (XGBoost); (c) conventional data-driven paradigm (RF).}
\label{Figure9}
\end{figure}

Fig.\ref{Figure9} further presents the susceptibility mapping results for the Qilian Permafrost Region, Tibetan Plateau study area. To maintain evaluative consistency and objectivity throughout the study, we again extracted the Jenks Natural Breaks from our paradigm's full-dataset predictions. These thresholds, after smooth rounding (0.04, 0.12, 0.22, and 0.36), serve as the unified spatial hazard classification baseline for this region.

The Qilian Permafrost Region, Tibetan Plateau study area is characterized by extreme topographic relief: the northern section comprises relatively gentle, high-altitude terraces, whereas the central and southern sectors feature intensely incised valleys and steep V-shaped gorges. Within this highly complex geomorphic environment, spatial discrepancies among the predictive paradigms become markedly more pronounced.

Regarding spatial distribution, our proposed paradigm maintains exceptional physical robustness and geomorphic coherence. The identified very high (red) and high (orange) susceptibility zones are accurately and contiguously anchored along the flanks of the deeply incised gorges and dendritic drainage networks in the central and southern sectors. Simultaneously, for the gentle northern terraces, the model yields highly constrained and accurate very low-risk (dark blue) predictions. This spatial signature---continuous high-risk belts along valleys contrasted with expansive, homogenous low-risk areas---demonstrates that the integrated geomorphic prior knowledge effectively guides the model in capturing macroscopic topographic controls.

Conversely, purely data-driven models of the conventional data-driven paradigm suffer severe predictive distortions under these complex environmental conditions. Random Forest (RF) exhibits extreme spatial fragmentation. High-risk belts, which physically should distribute continuously along the incised gorges, shatter into irregular ``salt-and-pepper'' noise in the RF landslide susceptibility map. The abrupt juxtaposition of dark red and dark blue patches completely disrupts the spatial continuity of the gorge geomorphology, further corroborating RF's critical overfitting flaw tied to local feature artifacts.

Although XGBoost partially mitigates the extreme noise of RF via its internal regularization mechanisms---broadly outlining the high-risk zones (orange) across the central and southern gorges---it still displays distinct ``patchy'' characteristics when delineating the core very high-risk zones (red). Lacking guidance from physical mechanisms, it can only approximate hazard intervals mathematically, failing to precisely isolate and trace the continuous geophysical boundaries controlling slope instability as effectively as the proposed paradigm.

In summary, the results from the Qilian Permafrost Region, Tibetan Plateau study area consistently demonstrate that the proposed paradigm delivers stable, efficient, and spatially coherent landslide susceptibility predictions under real-world conditions of severe class imbalance, incomplete environmental factors, and extreme data scarcity---substantiating both the practical utility and the strong adaptability of the proposed paradigm to complex real-world application scenarios.

\section{Discussion}

\subsection{Physical information embedded in topography: A long-underestimated source of prior knowledge}

Topography is not a random collection of geometric shapes, but rather a physical record of the interplay among tectonic uplift, climatic erosion, and gravity-driven mass movement over geological timescales. Topographic parameters such as slope gradient and upstream contributing area are not merely geometric metrics; they serve as proxies for the driving stress, hydrological convergence, and erosion rates acting on hillslopes. In strength-limited geomorphic regions, where slope angles and bedrock exposure evolve with increasing erosion rates, topographic form directly reflects slope stability thresholds \citep{RN29}. However, conventional data-driven paradigm reduces topographic derivatives --- slope, curvature, and related indices --- to isolated statistical predictors, thereby failing to capture the underlying geomechanical constraints encoded in terrain morphology. Under conditions of extreme sample scarcity, models are unable to infer reliable geomechanical relationships from scarce data alone, and predictions frequently deviate from geomorphologically consistent patterns \citep{RN26}.

The core insight of this study is that topography inherently constitutes an objective source of slope stability prior knowledge, one that is independent of historical landslide inventories. The joint distribution of average drainage area slope and drainage area physically encodes the relative magnitude of driving stresses across hillslopes within a catchment \citep{RN14}. Incorporating these physically derived constraints as explicit geomorphic prior knowledge establishes physically plausible boundaries for subsequent statistical inference. As a result, even a minimal number of labeled landslide samples can be utilized effectively, and the unconstrained overfitting that commonly afflicts purely data-driven models under data-scarce conditions is avoided.

Results from this study show that geomorphic prior knowledge yields greater performance gains under data-scarce conditions than under data-rich conditions (A detailed comparative analysis is provided in the Supporting Information B.). This finding corroborates the proposed mechanism: when observational samples are limited, the physical information embedded in topography is most valuable, partially compensating for the deficit in statistical training signal caused by scarce data. This result carries an important implication for regional geohazard assessment: terrain analysis should be recognized as a primary source of physical prior knowledge in landslide susceptibility mapping, rather than treated as one predictor among many statistical predictors \citep{RN30}.

\subsection{Extracting statistical patterns from scarce landslide data and the selection of predictive models}

The value of geomorphic prior knowledge is acknowledged; however, the extraction of statistical patterns within scarce landslide data should not be abandoned. Even under conditions of extreme sample scarcity, actual landslide occurrence patterns within specific geological environments are still reflected by scarce historical landslide records. The core challenge lies in how these patterns can be effectively extracted and noise can be suppressed under the data scarce conditions.

Conventional data-driven paradigm depends heavily on large landslide samples. When the sample size drops sharply, systematic biases are easily generated by the accidental distribution of limited data \citep{RN26}. Therefore, the tabular foundation model (TFM) was selected as the core inference engine in the proposed paradigm. Through offline pre-training on massive synthetic datasets, universal statistical patterns are internalized into the model parameters. When scarce landslide data are processed, complex probability distributions do not need to be estimated from scratch by the TFM. Instead, the scarce samples are treated as local evidence to perform Bayesian updates on the universal prior. Thus, stable probability inferences can still be provided, and the overfitting problem of conventional models under data-scarce conditions is effectively avoided \citep{RN20} (A detailed comparative analysis of TFM performance is provided in the Supporting Information C.).

It should be noted that the theoretical contribution of the proposed paradigm is not bound to the TFM algorithm specifically. Rather, the central contribution is the dual-driven architecture that combines geomorphic prior knowledge with scarce landslide data. Any algorithm that maintains robust performance under small-sample conditions --- including regularization-based methods such as ridge regression or elastic net --- is equally applicable in the statistical inference role. Within this paradigm, geomorphic prior knowledge defines the global distribution of mechanically unstable terrain, while the small-sample learning model refines this baseline by identifying high-risk spatial patterns specific to the local geological environment. The synergy between these two information sources is what enables the proposed paradigm to sustain high predictive accuracy under conditions of dual-scarce.

\subsection{Implications for geohazard risk assessment and mitigation}

Significant implications for both theoretical and engineering applications are provided by the findings of this study. At the theoretical level, empirical evidence is provided that the geomechanical stability information encoded in topography is an operational and universal source of prior knowledge for geohazards. Its inherent value has been long underestimated in conventional data-driven paradigm. The gap between physically-based deterministic modeling and data-driven statistical modeling is bridged by the deep integration of physical knowledge and scarce-data \citep{RN31}. Consequently, researchers are urged to move beyond simple ``multi-factor statistical splicing'' and transition toward ``physics-guided spatial inference.''

At the engineering level, an accessible geohazard assessment solution is provided by this paradigm for data-scarce mountainous regions worldwide. In the Qilian Mountains permafrost region of the Tibetan Plateau ---a region characterized by severe sample imbalance and incomplete environmental conditioning factors --- geoscientifically coherent landslide susceptibility maps were successfully generated. These results suggest that the paradigm is applicable to many remote, high-risk geological regions lacking landslide inventories, including the Himalayan foothills, the Andes, and the East African Rift. In such regions, reliable susceptibility maps can be produced using only open-source digital elevation models and a small number of field-verified landslide sample data, keeping data acquisition requirements low and substantially improving the accessibility of high-accuracy hazard assessment in data-scarce environments.

\subsection{Limitations and future work}

Despite the demonstrated advantages of the proposed paradigm, two limitations of the proposed paradigm warrant acknowledgment. First, the geomorphic prior knowledge is grounded in Culmann slope stability mechanics, which is most applicable to shallow translational failures \citep{RN14}. Although results from this study confirm that geomorphic priors enhance predictive accuracy for composite landslide inventories, their effectiveness may be reduced in regions dominated by deep-seated rotational failures or complex multi-mechanism slope movements. The dominant failure mode of the study region should therefore be assessed before geomorphic prior-driven performance gains are interpreted. Second, geomorphic priors are computed here using a single-scale domain-wide joint cumulative distribution function. Compared to the multi-scale hierarchical strategy of the original morphometric formulation \citep{RN14}, this simplification may underestimate prior susceptibility in topographically heterogeneous subregions. Future implementations should explore adaptive multi-scale prior estimation to improve local spatial precision.

Beyond these limitations, several directions offer promising opportunities for extending the proposed paradigm. Incorporating time-varying geophysical signals --- such as InSAR-derived surface deformation data and outputs from freeze--thaw process models --- into the prior system would allow the framework to move from static susceptibility mapping toward temporally resolved risk forecasting. Systematic assessment of the paradigm's transferability across a broader range of geological environments would further consolidate its theoretical foundations for geohazard assessment in data-scarce regions. The core principle demonstrated here --- that terrain morphology encodes physically meaningful slope stability information capable of anchoring statistical learning under data scarcity --- provides a transferable conceptual basis for geohazard assessment in the many regions of the world where conventional data-intensive approaches remain impractical.

\section{Conclusion}

This study addresses the pervasive challenge of data scarcity in geohazard assessment across global mountainous and plateau regions, proposing novel knowledge-data dually driven paradigm for landslide susceptibility prediction that integrates geomorphic prior knowledge with scarce landslide data. Through verification in the data-rich Central Italy study area and subsequent practical application in the data-scarce Qilian Permafrost Region, Tibetan Plateau study area, the principal conclusions are drawn as follows:

First, the proposed paradigm substantially reduces data requirements without compromising predictive reliability. In the Central Italy, where results obtained by conventional data-driven paradigm trained on the full available landslide data served as the performance baseline, the proposed paradigm achieved equivalent predictive accuracy using approximately 30\% of available landslide data. This finding demonstrates that geomorphic prior knowledge functions as a powerful physical constraint that compensates for the information loss associated with drastically reduced sample sizes, while the small-sample inference capability of the tabular foundation model efficiently extracts spatial patterns of landslide occurrence from scarce landslide data. Together, these mechanisms fundamentally reduce the paradigm's dependence on large-scale, costly historical landslide data.

Second, the paradigm exhibits robust predictive performance under genuinely data scarcity. In the Qilian Mountains permafrost region of the Tibetan Plateau --- characterized by incomplete environmental conditioning factors, severe landslide-to-non-landslide class imbalance, and limited total observations --- the proposed paradigm successfully generated susceptibility maps with high geomorphic spatial coherence. This result confirms that terrain-derived geomorphic priors, which encode the mechanical instability signatures of the landscape, can stabilize prediction under data-scarce conditions.

Third, the proposed paradigm provides an accessible solution for geohazard risk assessment under globally data-scarce conditions. In practical applications, this paradigm requires only an open-source digital elevation model (to extract geomorphic priors) and a minimal amount of landslide sample data to achieve highly reliable spatial predictions of landslide susceptibility. Such minimal data dependency makes the paradigm highly applicable to under-surveyed and ecologically fragile regions, such as the Hengduan Mountains, the Himalayas, and the Andes. It enables the rapid and cost-effective generation of reliable landslide susceptibility maps, significantly reducing the burden of extensive field investigation.

Despite these contributions, two limitations of the current study should be noted. The geomorphic prior knowledge is grounded in shallow translational slope failure mechanics, and its indicative value may diminish in regions dominated by deep-seated rotational landslides. Additionally, the single-scale domain-wide joint cumulative distribution function computation adopted here may introduce local bias in topographically heterogeneous regions relative to the multi-scale hierarchical strategy of the original morphometric formulation. Future research should prioritize the integration of time-varying geophysical signals (including InSAR-derived surface deformation and freeze--thaw cycle dynamics) into the prior system to extend the framework toward temporally resolved risk forecasting, and should assess the paradigm's transferability across a broader range of geological environments to further consolidate the theoretical foundations of geohazard assessment in data-scarce regions.

\bibliographystyle{elsarticle-harv}
\bibliography{yang_reference}

\end{document}